\pdfoutput=1
\documentclass[letterpaper, 10 pt, journal, twoside]{IEEEtran}

\usepackage{times}
\usepackage[pdftex]{graphicx}
\graphicspath{{figures/}}
\usepackage{amsmath,amssymb,amsfonts,bm}
\usepackage{mathtools}
\usepackage[space]{cite}
\usepackage{booktabs}
\usepackage{multirow}
\usepackage{array}
\usepackage{siunitx}
\usepackage{url}
\usepackage{xcolor}
\usepackage{soul}
\usepackage{hyperref}

\DeclareRobustCommand{\yellowtext}[1]{#1}

\newsavebox{\threedgutbox}

\hypersetup{colorlinks=true, linkcolor=black, citecolor=black, urlcolor=black}
\IEEEoverridecommandlockouts
\makeatletter
\def\section{\@startsection{section}{1}{\z@}{2.4ex plus 0.8ex minus 0.8ex}%
{0.6ex plus .3ex minus 0ex}{\normalfont\normalsize\scshape\centering}}%
\def\subsection{\@startsection{subsection}{2}{\z@}{1.9ex plus 0.7ex minus 0.7ex}%
{0.4ex plus .3ex minus 0ex}{\normalfont\normalsize\itshape}}%
\def\subsubsection{\@startsection{subsubsection}{3}{\parindent}{0.3ex plus 0.1ex minus 0.1ex}%
{0ex}{\normalfont\normalsize\itshape}}%
\makeatother

\makeatletter
\let\origmaketitle\maketitle
\renewcommand{\maketitle}{\begingroup\let\Huge\LARGE\origmaketitle\endgroup}
\makeatother

\newcommand{\SO}{\mathrm{SO}(3)}

\newcommand{\R}{\mathbb{R}}
\newcommand{\Exp}{\mathrm{Exp}}
\newcommand{\Log}{\mathrm{Log}}
\newcommand{\Skew}[1]{\left[#1\right]_{\times}}
\newcommand{\Norm}[1]{\left\lVert #1 \right\rVert}

\title{Rolling-Shutter Aware LiDAR-Inertial-Visual Odometry}

\author{Haoda Zhu$^{1,{\dagger}}$, Xiangyu Hong$^{2,{\dagger}}$, Chunran Zheng$^{1}$, Haotian Li$^{2,*}$, and Fu Zhang$^{1,*}$%
\thanks{Manuscript received: May 26, 2026; Revised: July 26, 2026; Accepted: October 4, 2026.}%
\thanks{This paper was recommended for publication by Editor Javier Civera upon evaluation of the Associate Editor and Reviewers comments. This work was supported by Spatial Robotics Limited.}%
\thanks{$^{1}$Haoda Zhu, Chunran Zheng, and Fu Zhang are with the
Department of Mechanical Engineering, The University of Hong Kong, Hong Kong SAR, China.
{\tt\footnotesize haodazhumars@connect.hku.hk, zhengcr@connect.hku.hk, fuzhang@hku.hk}}%
\thanks{$^{2}$Xiangyu Hong and Haotian Li are with Spatial Robotics Limited, Hong Kong SAR, China.
{\tt\footnotesize eric.hong@sparo.hk, cyrus.li@sparo.hk}}%
\thanks{$^{\dagger}$Haoda Zhu and Xiangyu Hong contributed equally to this work. $^{*}$Corresponding authors: Haotian Li and Fu Zhang.\hfil\break
{\tt\footnotesize cyrus.li@sparo.hk, fuzhang@hku.hk}}%
\thanks{Digital Object Identifier (DOI): see top of this page.}%
}

\IEEEaftertitletext{\vspace*{-1.5\baselineskip}}

\begin{document}
\maketitle
\begin{abstract}
This paper presents FAST-LIVO-RS, a rolling-shutter~(RS) aware direct LiDAR-inertial-visual odometry system that addresses the row-dependent geometric inconsistency arising when RS images are processed under global-shutter~(GS) assumptions. In direct LIVO pipelines, this inconsistency degrades photometric optimization, point-cloud colorization, and robustness under fast motion. To capture intra-frame motion over the RS readout time, FAST-LIVO-RS introduces adaptive state augmentation within a sequential error-state iterated Kalman filter (ESIKF), dynamically inserting clone states during IMU propagation based on estimated angular and linear acceleration to provide a motion-adaptive piecewise-linear trajectory approximation. Building on this representation, a row-consistent direct visual update is formulated through fixed-point row-time projection and Jacobian distribution to clone-segment endpoints, ensuring that each photometric residual is linearized at the correct row-specific camera pose. RS-consistent visual map management keeps patch maintenance and reference-frame updates consistent with row-specific RS geometry rather than single end pose assumptions. We evaluate FAST-LIVO-RS on a self-collected RS dataset and four GrandTour tasks, together with a comparison using synchronized RS and GS cameras and 3D reconstruction. The results show that row-time modeling improves robustness on RS sequences and provides localization and reconstruction quality comparable to FAST-LIVO2 using the synchronized GS camera.
\end{abstract}

\vspace{-5pt}
\section{Introduction}
\IEEEPARstart{L}{iDAR}-inertial-visual odometry (LIVO) has become an important solution for robotic navigation and mapping in GPS-denied environments. By tightly combining LiDAR geometry, inertial dynamics, and visual texture, LIVO systems provide strong robustness and accuracy under complex motions and scenes. Representative systems such as FAST-LIVO \cite{zheng2022fastlivo}, R$^3$LIVE \cite{lin2021r3live}, and FAST-LIVO2 \cite{zheng2024livo2} have demonstrated that direct frame-to-map fusion within iterated Kalman filtering frameworks enables efficient onboard deployment. In particular, by reusing LiDAR map points as visual anchors and performing direct photometric alignment without feature extraction, these systems achieve state-of-the-art accuracy across diverse scene types.

Despite this progress, most existing LIVO systems are built around global-shutter (GS) camera models. In practice, rolling-shutter (RS) cameras are highly attractive for robotics: at similar hardware cost, RS CMOS sensors offer higher resolution, higher frame rates, better signal-to-noise ratio, wider dynamic range, and lower power consumption than GS alternatives. RS cameras also dominate commodity smartphones and industrial platforms, broadening practical deployment scenarios. The challenge is that RS sensors expose rows sequentially, so each row corresponds to a different camera pose. Under fast motion, this row-dependent imaging introduces nonlinear time-space coupling that invalidates the single-pose projection assumption, propagating errors to visual residuals, patch warping, point-cloud colorization, and image undistortion.

In the VO/VIO literature, several RS aware methods have been proposed, built on direct photometric optimization or on continuous-time motion representations. However, they are developed without LiDAR geometric constraints, so directly transplanting them to a LIVO system is not straightforward: tightly coupled filter updates, LiDAR-supported visual maps, and real-time onboard execution all require mechanisms that existing RS designs do not provide.

To address this gap, we develop FAST-LIVO-RS, a rolling-shutter aware direct LIVO system that keeps the sequential LiDAR-then-visual fusion paradigm while introducing RS-consistent modeling across state estimation and visual map management. The main contributions are:
\begin{itemize}
    \item \textbf{Adaptive state augmentation for RS motion}. We adaptively augment the ESIKF state with pose clones during IMU propagation to capture intra-frame motion nonlinearity, while preserving consistent covariance coupling between base and augmented states.
    \item \textbf{Row-consistent direct visual update}. We formulate RS aware residual construction with fixed-point row-time projection, distributing Jacobians to clone-segment endpoints for stable direct photometric updates under row-dependent camera poses.
    \item \textbf{RS-consistent visual map management}. We keep visual map retrieval, patch maintenance, and reference-frame updates consistent with row-specific RS geometry rather than single end pose assumptions.
    \item \textbf{Downstream reconstruction}. We evaluate the effectiveness of FAST-LIVO-RS through three downstream uses: point-cloud colorization, RS image undistortion, and reconstruction with a fixed 3DGUT~\cite{wu20253dgut} backend.
\end{itemize}

\begin{figure*}[t]
    \centering
    \includegraphics[width=0.96\textwidth]{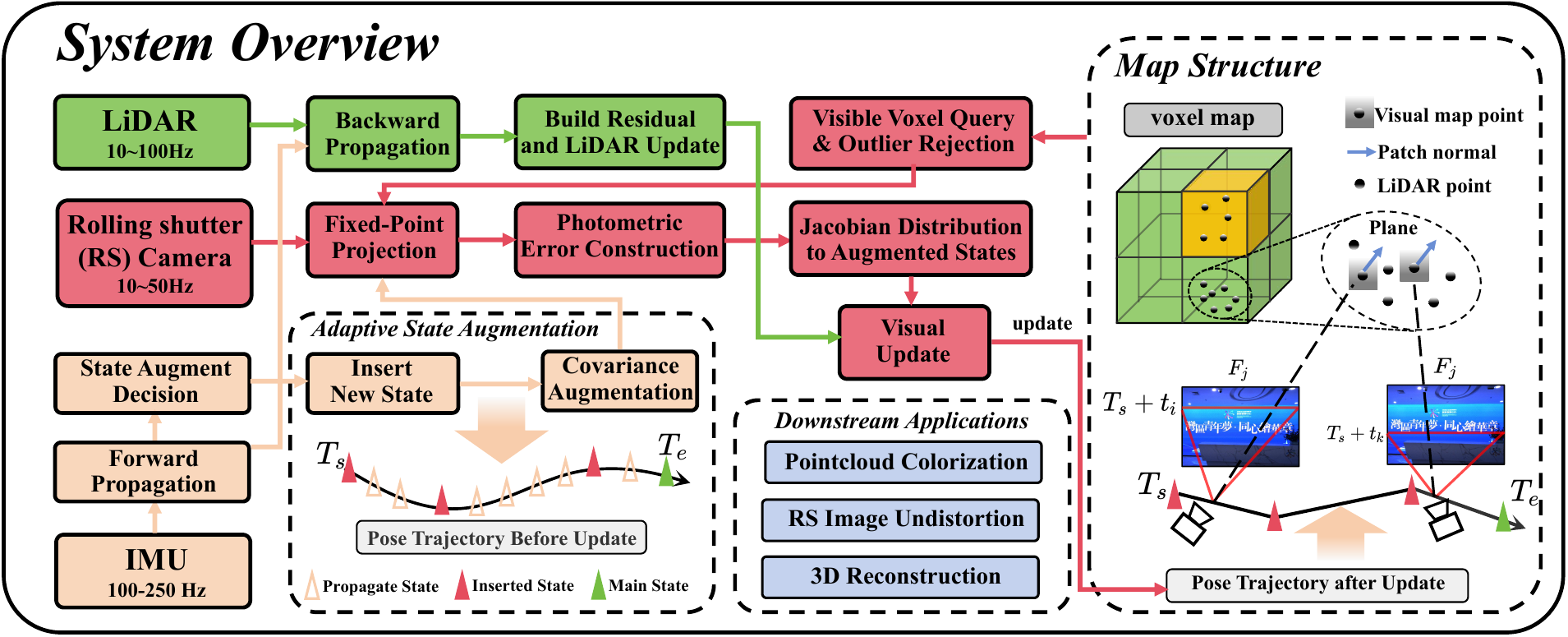}
    \caption{System overview of FAST-LIVO-RS.}
    \label{fig:system}
\end{figure*}
\section{Related Work}
\subsection{LiDAR-Inertial-Visual Odometry}
The fusion of LiDAR, camera, and IMU measurements overcomes degeneration modes of standalone sensors: geometry-degenerate corridors challenge LiDAR-only systems~\cite{zhang2014loam,xu2021fast1,xu2022fast2}, while low-texture regions degrade visual pipelines~\cite{VINSmono}. LVI-SAM~\cite{shan2020lvisam} achieves tight coupling through factor-graph fusion with mutually supporting LIO and VIO subsystems. Within this broader family, LIC-Fusion and Super Odometry use discrete-time estimation backends~\cite{zuo2019licfusion,zhao2021superodom}, whereas CLIC and Coco-LIC represent motion continuously \cite{lv2023clic,lang2023cocolic}. R$^3$LIVE~\cite{lin2021r3live} introduces direct photometric updates within an on-manifold iterated Kalman filter, enabling RGB-colored mapping without feature extraction. FAST-LIVO~\cite{zheng2022fastlivo} introduces measurement-level tight coupling by reusing LiDAR map points as visual anchors, and FAST-LIVO2~\cite{zheng2024livo2} advances this line through a unified voxel map, sequential ESIKF updates, and dynamic reference patch management.

The direct LIVO systems summarized above assume a single GS pose for all rows. FAST-LIVO-RS instead extends direct LIVO with row-consistent photometric updates and visual-map management.

\subsection{Rolling-Shutter Visual Odometry}
Early work established the non-central projection model of RS cameras~\cite{meingast2005geometric}, and RS constraints were incorporated into visual bundle adjustment~\cite{hedborg2012rsba}. Li et al. couple inertial sensing with an RS camera model in an EKF~\cite{li2013rsvio}. DSO-RS~\cite{schubert2018dsors} extends direct sparse odometry with constant-velocity RS modeling, showing that neglecting RS effects can cause large drift. Schubert~\textit{et al.}~\cite{schubert2019rsvio} couple constant-twist motion with IMU measurements for metric-scale RS-VIO, while Ctrl-VIO~\cite{lang2022ctrlvio} uses uniform cubic B-splines to represent continuous-time RS motion beyond the constant-velocity regime. However, its fixed knot spacing does not adapt to actual motion dynamics and can introduce redundant temporal resolution during slow motion.

Direct RS estimation has also been studied in semi-dense SLAM~\cite{kim2016rsslam} and modern RS bundle adjustment~\cite{liao2023rsba}. Compared with compact constant-velocity and constant-twist models and fixed-spacing spline trajectories, FAST-LIVO-RS uses estimated angular and linear acceleration to adaptively insert a bounded number of covariance-coupled intra-frame clones during IMU propagation in a sequential ESIKF, concentrating temporal support in portions of the readout interval with stronger motion variation and directly applying the resulting row-time poses to LiDAR-supported photometric residuals.

\section{System Overview}
\label{sec:overview}

Fig.~\ref{fig:system} illustrates the architecture of FAST-LIVO-RS. The system keeps the sequential ESIKF pipeline \cite{zheng2024livo2}, while adding adaptive state augmentation for RS motion, a row-consistent direct visual update, and RS-consistent visual map management. LiDAR points are first undistorted to the scan end and fused through geometric residuals; the RS camera branch then constructs row-consistent photometric residuals and updates both the base state and active augmented states. Each visual map point stores the row-specific reference geometry derived from the trajectory at its creation time, so later patch matching, point-cloud colorization, image undistortion, and 3DGUT-based reconstruction use the same RS-consistent geometry.

\section{Method}
\label{sec:method}

\subsection{State Representation and IMU Propagation}
\label{sec:state_propagation}

We adopt the manifold-based $\boxplus / \boxminus$ notation from~\cite{zheng2024livo2}. The IMU frame~$I$ serves as the body frame and the first body frame defines the global frame~$G$. The base state resides on the compound manifold $\mathcal{M} \triangleq \SO \times \R^{16}$:
\begin{equation}
    \mathbf{x} \triangleq \bigl[{}^G\mathbf{R}_I^{\!\top},\; {}^G\mathbf{p}_I^{\!\top},\; {}^G\mathbf{v}_I^{\!\top},\; \mathbf{b}_g^{\!\top},\; \mathbf{b}_a^{\!\top},\; {}^G\mathbf{g}^{\!\top},\; \tau\bigr]^{\!\top} \in \mathcal{M},
    \label{eq:state}
\end{equation}
where ${}^G\mathbf{R}_I \in \SO$, ${}^G\mathbf{p}_I \in \R^3$, and ${}^G\mathbf{v}_I \in \R^3$ denote the world-frame IMU attitude, position, and velocity, $\mathbf{b}_g, \mathbf{b}_a \in \R^3$ are the gyroscope and accelerometer biases, ${}^G\mathbf{g}$ is the gravity vector, and $\tau$ is the inverse camera exposure parameter~\cite{zheng2024livo2}. To capture intra-frame RS motion, the state is augmented with up to $K_{\max}$ pose clones:
\begin{equation}
    \mathbf{x}_{\mathrm{aug}} \triangleq \bigl[\mathbf{x}^{\!\top},\; {}^G\mathbf{R}_{I,0}^{\!\top},\; {}^G\mathbf{p}_{I,0}^{\!\top},\; \ldots,\; {}^G\mathbf{R}_{I,K-1}^{\!\top},\; {}^G\mathbf{p}_{I,K-1}^{\!\top}\bigr]^{\!\top},
    \label{eq:state_aug}
\end{equation}
where $({}^G\mathbf{R}_{I,k}, {}^G\mathbf{p}_{I,k})$ is the $k$-th clone of the world-frame IMU pose, yielding error-state dimension $n = \dim(\mathcal{M}) + 6K$. The base state and its error are related through
\begin{equation}
    \mathbf{x} = \hat{\mathbf{x}} \boxplus \delta\mathbf{x},
    \label{eq:boxplus}
\end{equation}
which acts as $\mathbf{R} = \hat{\mathbf{R}}\,\Exp(\delta\boldsymbol{\theta})$ on the $\SO$ component and as standard vector addition on all $\R^n$ components. IMU propagation operates only on $\mathbf{x}$; clone poses are frozen once inserted. The extrinsic $({}^C\mathbf{R}_I, {}^C\mathbf{p}_I)$ is pre-calibrated and fixed.

Upon receiving an IMU measurement $\mathbf{u} = [\boldsymbol{\omega}_m^{\!\top},\; \mathbf{a}_m^{\!\top}]^{\!\top}$ with associated noise covariance $\mathbf{Q}$, the forward propagation follows the standard discrete kinematics
\begin{equation}
    \hat{\mathbf{x}}_{i+1} = \hat{\mathbf{x}}_i \boxplus \bigl(\Delta t \cdot \mathbf{f}(\hat{\mathbf{x}}_i, \mathbf{u}_i, \mathbf{0})\bigr),
    \label{eq:propagation}
\end{equation}
where $\Delta t$ is the sampling interval. The specific forms of the dynamics $\mathbf{f}(\cdot)$ and the Jacobians $\mathbf{F}_{\delta\mathbf{x}}$, $\mathbf{F}_\mathbf{w}$ follow~\cite{zheng2024livo2}. The error-state covariance propagates as
\begin{equation}
    \hat{\mathbf{P}}_{i+1} = \mathbf{F}_{\delta\mathbf{x}}\,\hat{\mathbf{P}}_i\,\mathbf{F}_{\delta\mathbf{x}}^{\!\top} + \mathbf{F}_\mathbf{w}\,\mathbf{Q}\,\mathbf{F}_\mathbf{w}^{\!\top}.
    \label{eq:cov_prop}
\end{equation}

\subsection{LiDAR Update}
\label{sec:lidar}

Raw LiDAR points are undistorted to scan-end time using IMU-propagated poses~\cite{zheng2024livo2}. For each correspondence~$i$, the point-to-plane residual is
\begin{equation}
    r_i = \mathbf{n}_i^{\top}\bigl({}^G\mathbf{R}_I\,\mathbf{p}_i^b + {}^G\mathbf{p}_I\bigr) + d_i,
    \label{eq:lidar_res}
\end{equation}
where $\mathbf{n}_i \in \R^3$ is the unit plane normal, $d_i$ the signed offset, and $\mathbf{p}_i^b$ the undistorted body-frame point. The standard ESIKF LiDAR update~\cite{zheng2024livo2}, whose residual depends only on the end pose, provides the prior for the subsequent visual update.

\subsection{Rolling-Shutter Timing and Adaptive State Augmentation}
\label{sec:cloning}

\begin{figure}[t]
    \centering
    \includegraphics[width=0.43\textwidth]{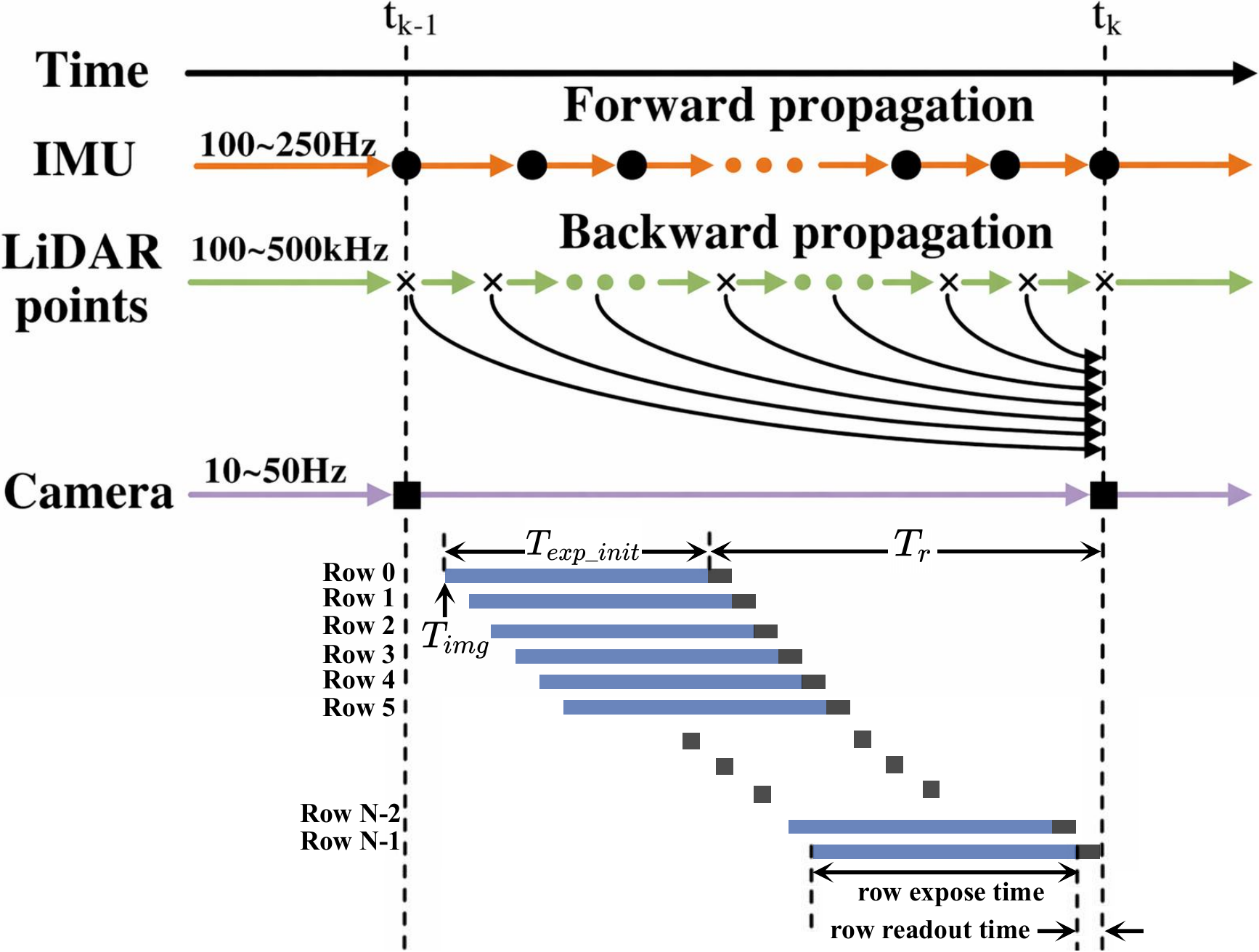}
    \caption{Multi-sensor timing model for IMU propagation, LiDAR point undistortion, and RS row exposure.}
    \label{fig:timing}
\end{figure}

Unlike GS cameras that capture all rows simultaneously, an RS sensor exposes each row $v \in [0, H\!-\!1]$ at a distinct time over a readout time of duration $T_r$. Two calibrated timing parameters define this interval: $T_r$, the full-frame readout duration, and $T_{\exp\_init}$, the constant timing offset from the camera driver timestamp \ensuremath{T_{img}} to the first-row exposure timestamp. The end-row and start-row times are
\begin{equation}
    T_e = T_{img} + T_{\exp\_init} + T_r, \qquad T_s = T_e - T_r,
    \label{eq:timing}
\end{equation}
and the normalized row time
\begin{equation}
    s(v) = \frac{v}{H - 1} \in [0,1]
    \label{eq:sv}
\end{equation}
  maps row~$0$ to $T_s$ and row~$H\!-\!1$ to $T_e$ (Fig.~\ref{fig:timing}). Equivalently, row~$v$ is exposed
  at $T(v)=T_s+s(v)T_r$. This row-time parameterization defines the readout interval over which pose clones
  are inserted and later interpolated, allowing RS visual residuals to be evaluated at row-specific camera
  poses rather than at a single frame pose.
\subsubsection{Interpolation Error Bound}
Within the sequential ESIKF, IMU-derived motion variation determines where the bounded intra-frame clone support is allocated. Adaptive state augmentation concentrates temporal resolution where motion deviates most from linear interpolation. Since the interpolation error over a sub-interval of length $\Delta t$ scales as $\tfrac{1}{8}\Norm{\ddot{\mathbf{q}}}\Delta t^2$, we bound the clone spacing using IMU-estimated angular acceleration and gravity-compensated acceleration:
\begin{align}
    \dot{\boldsymbol{\omega}}_i = \frac{\boldsymbol{\omega}_i - \boldsymbol{\omega}_{i-1}}{\Delta t_i},
    \quad \boldsymbol{\omega}_i = \boldsymbol{\omega}_{m,i} - \hat{\mathbf{b}}_g, \label{eq:wdot}\\
    \hat{\mathbf{a}}_i = {}^G\hat{\mathbf{R}}_I(\mathbf{a}_{m,i} - \hat{\mathbf{b}}_a) + {}^G\hat{\mathbf{g}}, \label{eq:accel}\\
    \Delta t_{lim}^{\omega} = \max\!\left(\Delta t_{min},\; \sqrt{\frac{8\,\theta_{tol}}{\Norm{\dot{\boldsymbol{\omega}}_i} + \epsilon}}\right).
    \label{eq:dtlim_w}\\
    \Delta t_{lim}^{a} = \max\!\left(\Delta t_{min},\; \sqrt{\frac{8\,d_{tol}}{\Norm{\hat{\mathbf{a}}_i} + \epsilon}}\right).
    \label{eq:dtlim_a}\\
    \Delta t_{lim} = \min\!\bigl(\Delta t_{lim}^{\omega},\; \Delta t_{lim}^{a}\bigr),
    \label{eq:dtlim}
\end{align}
where $\epsilon$ avoids division by zero and $\Delta t_{min}$ prevents numerically redundant adjacent clones. A new clone is inserted whenever the elapsed time since the last insertion reaches $\Delta t_{lim}$, provided that the clone budget $K < K_{\max}$ is not exhausted.

\subsubsection{Clone Insertion and Covariance Augmentation}
The first clone is always placed at $T_s$. When the IMU integration step straddles $T_s$, the integrator propagates the state to $T_s$, records clone~$0$, and then continues to the end of the original step. When clone~$k$ is inserted, its pose $({}^G\mathbf{R}_{I,k}, {}^G\mathbf{p}_{I,k})$ copies the current propagated values. The terminal knot $s_K = 1$ is assigned to the end pose $({}^G\mathbf{R}_I, {}^G\mathbf{p}_I)$ after propagation to $T_e$, so the knot vector $\mathbf{s} = \{s_0, \ldots, s_K\}$ always spans the full readout time. Retaining at most $K_{\max}$ clones bounds the augmented state dimension while the terminal base pose completes the trajectory support over the readout interval.

Under the first-order error-state approximation, the clone is an exact copy of the current end pose. The selection Jacobian that maps the current error state to the new clone error is
\begin{equation}
    \mathbf{J} = \begin{bmatrix}
    \mathbf{I}_3 & \mathbf{0}_{3\times3} & \mathbf{0}_{3\times(n-6)} \\
    \mathbf{0}_{3\times3} & \mathbf{I}_3 & \mathbf{0}_{3\times(n-6)}
    \end{bmatrix},
    \label{eq:sel_jac}
\end{equation}
where the two $3\times3$ identity blocks select the attitude and position components of the end pose. The augmented covariance $\mathbf{P}^{+} \in \R^{(n+6)\times(n+6)}$ is
\begin{equation}
    \mathbf{P}^{+} = \begin{bmatrix} \mathbf{I}_n \\ \mathbf{J} \end{bmatrix} \mathbf{P} \begin{bmatrix} \mathbf{I}_n \\ \mathbf{J} \end{bmatrix}^{\!\top}
    = \begin{bmatrix} \mathbf{P} & \mathbf{P}\mathbf{J}^{\!\top} \\ \mathbf{J}\mathbf{P} & \mathbf{J}\mathbf{P}\mathbf{J}^{\!\top} \end{bmatrix}.
    \label{eq:cov_aug}
\end{equation}
Because $\mathbf{J}$ consists only of identity sub-blocks, the entire augmentation reduces to block extraction and requires no matrix multiplication. The resulting covariance-coupled clones remain part of the sequential ESIKF during propagation and the joint base/clone photometric update.

\subsubsection{Covariance Propagation with Clones}
After insertion, clone nominals are frozen. Partitioning the augmented covariance into base-state ($\mathbf{P}_{II}$), clone ($\mathbf{P}_{CC}$), and cross ($\mathbf{P}_{IC}$) blocks, a single propagation step yields
\begin{equation}
    \mathbf{P}_{i+1} = \begin{bmatrix}
    \mathbf{F}_{\delta\mathbf{x}}\,\mathbf{P}_{II}\,\mathbf{F}_{\delta\mathbf{x}}^{\!\top} + \mathbf{F}_\mathbf{w}\,\mathbf{Q}\,\mathbf{F}_\mathbf{w}^{\!\top}  & \mathbf{F}_{\delta\mathbf{x}}\,\mathbf{P}_{IC} \\[4pt]
    \mathbf{P}_{IC}^{\!\top}\mathbf{F}_{\delta\mathbf{x}}^{\!\top}  & \mathbf{P}_{CC}
    \end{bmatrix}.
    \label{eq:cov_prop_clone}
\end{equation}
The cross-covariance $\mathbf{P}_{IC}$ is left-multiplied by $\mathbf{F}_{\delta\mathbf{x}}$ to reflect the growing divergence between frozen clones and the evolving end pose, while the clone self-covariance $\mathbf{P}_{CC}$ remains unchanged because clone nominals do not participate in the kinematic integration. Through this coupling, the LiDAR update also implicitly constrains the clone poses via the Kalman gain, reinforcing geometric consistency across both sensor modalities, even though the LiDAR residual itself depends only on the end pose. After the visual update converges, all clone variables are marginalized by discarding the corresponding rows and columns, resetting the augmented state back to the base-state manifold $\mathcal{M} = \SO \times \R^{16}$ before the next frame's propagation begins.

\subsection{Rolling-Shutter Projection and Pose Interpolation}
\label{sec:rs_projection}

Projecting a world point ${}^G\mathbf{p}_i$ onto an RS image requires the camera pose at the row where the projection lands, but the landing row depends on the projection result, creating a circular dependency known as the rolling-shutter constraint~\cite{meingast2005geometric, schubert2018dsors}. The row-consistent pixel $\mathbf{u}^* = (u_x^*, u_y^*)^{\top}$ satisfies
\begin{equation}
    \mathbf{u}^* = \pi\!\bigl({}^C\mathbf{T}_G(u_y^*) \;\; {}^G\mathbf{p}_i\bigr).
    \label{eq:rs_fp}
\end{equation}
In general, \eqref{eq:rs_fp} has no closed-form solution~\cite{meingast2005geometric}. Following~\cite{schubert2018dsors}, we solve it by fixed-point iteration: starting from the end row $v^{(0)} = H - 1$, each iteration evaluates the camera pose at $v^{(j)}$, projects ${}^G\mathbf{p}_i$, and updates to the vertical coordinate of the result. Let $v^*\triangleq u_y^*$ denote the converged image row. The active segment $m^*\in\{0,\ldots,K-1\}$ satisfies $s_{m^*}\leq s(v^*)\leq s_{m^*+1}$ and has local coordinate $\lambda^*=(s(v^*)-s_{m^*})/(s_{m^*+1}-s_{m^*})$. An interior knot is assigned to the preceding segment; hence the terminal row uses $m^*=K-1$ and $\lambda^*=1$. Knot $j<K$ denotes clone $j$, whereas knot $K$ is the terminal base pose. The procedure also returns the camera-frame coordinates ${}^C\mathbf{p}_i = {}^C\mathbf{R}_G^*\,{}^G\mathbf{p}_i + {}^C\mathbf{p}_G^*$.

The row-specific IMU pose is obtained via piecewise-linear interpolation over the trajectory knot vector. For normalized row time $s(v) \in [s_m, s_{m+1})$ with local parameter \ensuremath{\lambda = (s(v) - s_m)/(s_{m+1} - s_m)}, the interpolated attitude and position are
\begin{align}
    {}^G\mathbf{R}_I(v) &= {}^G\mathbf{R}_{I,m}\,\Exp\!\bigl(\ensuremath{\lambda}\,\Log({}^G\mathbf{R}_{I,m}^{\top}\,{}^G\mathbf{R}_{I,m+1})\bigr), \label{eq:Rinterp}\\
    {}^G\mathbf{p}_I(v) &= (1-\ensuremath{\lambda})\,{}^G\mathbf{p}_{I,m} + \ensuremath{\lambda}\,{}^G\mathbf{p}_{I,m+1}, \label{eq:pinterp}
\end{align}
The camera pose at row~$v$ follows from the pre-calibrated extrinsic:
\begin{align}
    {}^C\mathbf{R}_G(v) &= {}^C\mathbf{R}_I\,{}^G\mathbf{R}_I(v)^{\top}, \label{eq:Rcw}\\
    {}^C\mathbf{p}_G(v) &= -{}^C\mathbf{R}_G(v)\,{}^G\mathbf{p}_I(v) + {}^C\mathbf{p}_I. \label{eq:pcw}
\end{align}

\subsection{Row-Consistent Direct Visual Update}
\label{sec:visual_update}

The visual update uses the LiDAR-refined prior. Each visual map point ${}^G\mathbf{p}_i$ carries a planar patch with surface normal from the LiDAR voxel map. For each patch pixel~$q$ with offset $\Delta\mathbf{u}$ from the patch center $\mathbf{u}_i$, the photometric residual is
\begin{equation}
    r_q = \tau_k\,\mathbf{I}_k\!\bigl(\mathbf{u}_i + \Delta\mathbf{u}\bigr) - \tau_r\,\mathbf{I}_r\!\bigl(\mathbf{u}_i' + \mathbf{A}_i^{r}\,\Delta\mathbf{u}\bigr),
    \label{eq:residual}
\end{equation}
where $\tau_k, \tau_r$ are the exposure parameters of the current and reference frames~\cite{zheng2024livo2}, $\mathbf{I}_k$ and $\mathbf{I}_r$ are the corresponding image intensity functions, $\mathbf{A}_i^{r}$ is the plane-induced affine warp~\cite{zheng2024livo2}, $\mathbf{u}_i'$ is the reference-frame patch center, and the current-frame center $\mathbf{u}_i$ is obtained through the fixed-point row-time projection~\eqref{eq:rs_fp}. Here, $\Delta\mathbf{u}$ is a fixed sampling offset in the local patch template: a state update moves the projected patch center $\mathbf{u}_i$, while the relative sampling offsets remain unchanged.

\subsubsection{Jacobian Decomposition and Distribution}
The residual Jacobian with respect to the IMU pose at row $v^*$ is composed through the chain rule. Let
\begin{equation}
    \mathbf{J}_{img} = \tau_k \bigl[\partial \mathbf{I}_k/\partial u_x,\; \partial \mathbf{I}_k/\partial u_y\bigr]_{\mathbf{u}_i}, \quad
    \mathbf{J}_{\pi} = \frac{\partial\pi({}^C\mathbf{p}_i)}{\partial\,{}^C\mathbf{p}_i}
    \label{eq:Jimg_Jpi}
\end{equation}
denote the exposure-scaled $1\times 2$ image gradient and the $2\times 3$ projection Jacobian. The perturbation with respect to rotation and translation errors of the IMU pose at row~$v^*$ yields the row-pose Jacobians
\begin{align}
    \mathbf{J}_R^{row} &= \mathbf{J}_{img}\,\mathbf{J}_{\pi}\,\Skew{{}^C\mathbf{p}_i}, \label{eq:JR_row}\\
    \mathbf{J}_t^{row} &= -\mathbf{J}_{img}\,\mathbf{J}_{\pi}\,{}^C\mathbf{R}_G^*. \label{eq:Jt_row}
\end{align}
Because the row pose is the piecewise-linear interpolation of the bracketing knots $m^*$ and $m^*\!+\!1$ with weights \ensuremath{(1\!-\!\lambda^*,\,\lambda^*)}, the chain rule distributes the row-pose Jacobian to the two neighboring knots in the filter state:
\begin{align}
    \frac{\partial r_q}{\partial \delta\boldsymbol{\theta}_{m^*}} &= (1-\ensuremath{\lambda^*})\,\mathbf{J}_R^{row}, \quad
    \frac{\partial r_q}{\partial \delta\mathbf{p}_{m^*}} = (1-\ensuremath{\lambda^*})\,\mathbf{J}_t^{row}, \label{eq:H_end0}\\
    \frac{\partial r_q}{\partial \delta\boldsymbol{\theta}_{m^*+1}} &= \ensuremath{\lambda^*}\,\mathbf{J}_R^{row}, \quad\;\;\,
    \frac{\partial r_q}{\partial \delta\mathbf{p}_{m^*+1}} = \ensuremath{\lambda^*}\,\mathbf{J}_t^{row}, \label{eq:H_end1}
\end{align}
The factors $(1-\lambda^*,\lambda^*)$ are exact for linear position interpolation. For SLERP, they define a first-order endpoint approximation under small relative rotations between adjacent knots. This small-rotation condition specifies the operating regime of the endpoint approximation independently of the motion-dependent criterion used for temporal-support placement. Each photometric residual contributes at most two non-zero six-column pose blocks, so the pose-related part of the measurement Jacobian has local block sparsity. With $K_{\max}$, the augmented error-state dimension is bounded by $\dim(\mathcal{M})+6K_{\max}$.

\subsubsection{Iterated ESIKF Update}
At each outer ESIKF iteration, the current augmented state is first used to solve the fixed-point projection and determine the active interpolation segment for every valid patch. Within this conditional linearization, $v^*$, $m^*$, and $\lambda^*$ are held fixed while the photometric residuals and two-endpoint Jacobian blocks are evaluated and stacked. Unlike propagation, where clone nominals are frozen, the resulting augmented normal equations jointly update the base state and all active clone errors~\cite{ikf_gs}. The next outer iteration repeats the fixed-point projection and segment assignment at the updated state, thereby incorporating row changes and knot-boundary crossings through relinearization. After convergence, the refined base and clone poses define an updated piecewise-linear trajectory over $[T_s,T_e]$ that is consistent with the photometric evidence and the LiDAR--inertial prior.

\subsection{RS-Consistent Visual Map Management}
\label{sec:map_management}

As illustrated by the right-hand map-interaction branch of Fig.~\ref{fig:system}, each visual map point must preserve the row-specific reference geometry at creation. Under a GS assumption, the reference patch is warped using the single end pose of the creation frame. For RS cameras, however, a map point first detected at row $v_{feat}$ was imaged at camera pose ${}^C\mathbf{T}_G(v_{feat})$, which can differ from the end pose. This mismatch makes the stored reference patch inconsistent with the current row-specific projection, introducing biased photometric residuals and degrading map quality.

Accordingly, each map point stores (a)~the full creation-frame knot trajectory $\{({}^G\mathbf{R}_{I,k}, {}^G\mathbf{p}_{I,k}), s_k\}_{k=0}^{K}$, and (b)~the row-specific camera pose ${}^C\mathbf{T}_G(v_{feat})$ evaluated via~\eqref{eq:Rcw}--\eqref{eq:pcw}. The affine warp in~\eqref{eq:residual} uses this stored row-specific pose for the reference geometry, while visible-voxel query, outlier rejection, and new-point insertion use the same fixed-point row-time projection~\eqref{eq:rs_fp} as the visual update.

\subsection{Rolling-Shutter-Consistent Downstream Applications}
\label{sec:downstream}

Beyond state estimation, we evaluate FAST-LIVO-RS through three downstream applications: point-cloud colorization, RS image undistortion, and 3D reconstruction.

\subsubsection{Point-Cloud Colorization}
RGB assignment reuses the same RS projection as the visual update. For each LiDAR point selected for coloring, the fixed-point row refinement first determines the image row at which the point was observed. The interpolated pose from~\eqref{eq:Rinterp}--\eqref{eq:pcw} then projects the point into the RS image, and the converged pixel is sampled bilinearly if it lies inside the valid image domain. This avoids assigning colors from a camera pose that is inconsistent with the point's actual imaging row, which is especially important for high-contrast edges and distant textured structures.

\subsubsection{Image Undistortion}
Each RS frame is converted into a GS-equivalent image through a row homography
\begin{equation}
    \mathbf{H}_v = \mathbf{K}\,{}^C\mathbf{R}_I\,{}^G\mathbf{R}_I(s(v))^{\top}\,{}^G\mathbf{R}_I\,{}^C\mathbf{R}_I^{\top}\,\mathbf{K}^{-1},
    \label{eq:H_undist}
\end{equation}
where $\mathbf{K}$ is the intrinsic matrix, ${}^G\mathbf{R}_I$ is the end pose rotation, and ${}^G\mathbf{R}_I(s(v))$ is interpolated from~\eqref{eq:Rinterp}. The target image is defined at the end pose time $T_e$. For each target pixel, the source row is refined with the same fixed-point mechanism as~\eqref{eq:rs_fp}: the current row estimate defines $\mathbf{H}_v$, the homography proposes a source pixel, and the source row is updated from that pixel. We model the dominant rotational inter-row motion and leave pixels outside the valid RS image domain invalid, because no measured source pixel exists for them.

\subsubsection{\texorpdfstring{3D Reconstruction}{3D Reconstruction}}
3DGUT~\cite{wu20253dgut} extends 3DGS with support for rolling shutter. In this work, we keep the 3DGUT reconstruction backend fixed and use the row-time poses estimated by FAST-LIVO-RS as the camera-motion input, thereby enabling rolling-shutter-consistent 3D reconstruction.
\begin{figure}[t]
    \centering
    \includegraphics[width=0.94\columnwidth]{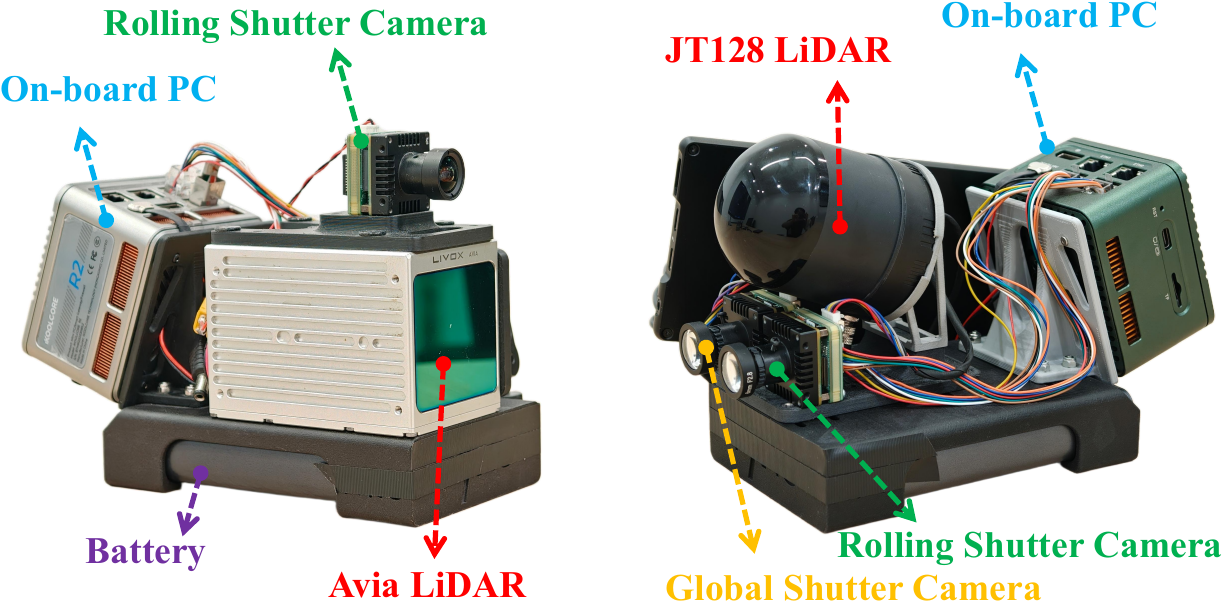}
    \caption{The two handheld sensor platforms used to collect the FAST-LIVO-RS dataset.}
    \label{fig:dataset_rigs}
\end{figure}
\begin{figure*}[t]
    \centering
    \includegraphics[width=0.99\textwidth]{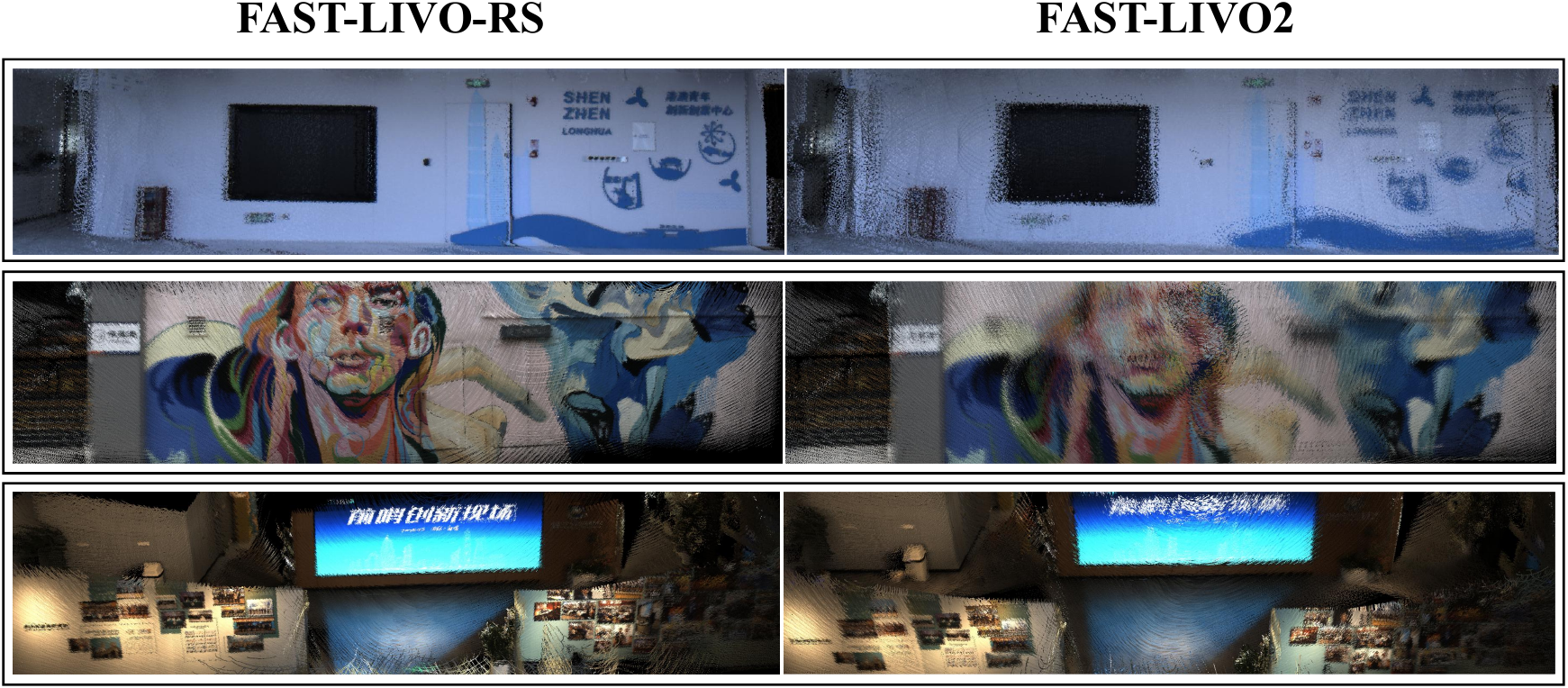}
    \vspace{-4pt}
    \caption{Point-cloud colorization comparison between FAST-LIVO-RS and FAST-LIVO2.}
    \label{fig:colorization_comparison}
    \vspace{-4pt}
\end{figure*}

\section{Dataset}
\label{sec:dataset}

We collect a LiDAR-inertial-RS dataset on two handheld platforms, denoted \emph{avia+rs} and \emph{jt+rs}, covering both end-to-end (E2E) sequences and motion-capture sequences, with deliberately aggressive motion to stress the adaptive state augmentation in Section~\ref{sec:cloning}.

\subsection{Readout Time Calibration}
The full-frame rolling-shutter readout duration $T_r$ is calibrated offline for the exact resolution, frame rate, and cropping mode used in our experiments, following the strobing-LED procedure provided by the open-source Gyroflow tool~\cite{gyroflow_rollingshutter}. The resulting $T_r$ is fixed for all sequences.

\subsection{Sensor Platform and Synchronization}
\label{sec:dataset_platform}

The two handheld rigs are shown in Fig.~\ref{fig:dataset_rigs}. The \emph{avia+rs} platform pairs a Livox Avia LiDAR with its built-in IMU and a 6\,MP industrial RS camera (MV-CB060-10GC) rigidly mounted on a handheld frame. Both platforms use the nominal calibrated readout time \ensuremath{T_r = 49.6\,\mathrm{ms}}. The \emph{jt+rs} platform uses a Hesai JT128 LiDAR with the same camera, plus a cost-matched 1.6\,MP GS camera (MV-CB016-10GC) as an auxiliary reference; this platform features a wider LiDAR field of view and validates compatibility with a pure solid-state LiDAR.

All sensors are hard-synchronized through a PTP-based hardware timing chain so that LiDAR, IMU, and camera measurements share a common clock. LiDAR-IMU temporal and extrinsic initialization uses LI-Init~\cite{zhu2022liinit}; LiDAR-camera extrinsic calibration uses FAST-Calib~\cite{zheng2025fastcalib}.

\subsection{Sequence Design and Scenarios}
\label{sec:dataset_sequences}

The dataset contains nine handheld sequences spanning indoor and outdoor scenes and two complementary evaluation modes. The \emph{jt+rs} platform provides six sequences, including two motion-capture recordings (JT-M1 and JT-M2), while the \emph{avia+rs} platform contributes three sequences collected with a distinct LiDAR scan pattern. The two motion-capture sequences provide high-precision pose ground truth for RMSE evaluation after timestamp alignment, while the remaining seven sequences are evaluated by E2E drift. The NOKOV motion-capture system used for JT-M1 and JT-M2 has a manufacturer-specified positional accuracy of \ensuremath{\pm0.5\,\mathrm{mm}}. Across three independent return-to-origin trials, the mean start-to-end displacement measured by this setup was \ensuremath{4.2\,\mathrm{mm}}. The dataset, together with calibrated extrinsics, will be publicly released upon acceptance.
\begin{table}[t]
    \centering
    \caption{RS model and solver settings.}
    \label{tab:rs_parameters}
    \resizebox{0.4\textwidth}{!}{%
    \begin{tabular}{l l}
        \toprule
        Parameter & Setting \\
        \midrule
        Maximum clone budget $K_{\max}$ & 4 \\
        Angular tolerance $\theta_{tol}$ & 0.002\,rad \\
        Translational tolerance $d_{tol}$ & 0.2\,cm \\
        Minimum interval $\Delta t_{min}$ & 0.005\,s \\
        Maximum fixed-point iterations $N_{fp}$ & 5 \\
        Maximum ESIKF iterations $N_{\mathrm{ESIKF}}$ & 10 per pyramid level \\
        \bottomrule
    \end{tabular}}
\end{table}

\begin{table*}[t]
    \centering
    \caption{Quantitative trajectory accuracy.\textsuperscript{1}}
    \label{tab:main_results}
    \resizebox{\textwidth}{!}{%
    \begin{tabular}{l l c c c c c c c c}
        \toprule
        Sequence & Scene/source & Duration [s] & Eval & Ours & Ours w/o adapt.\ aug. & Ours w/o RS map & FAST-LIVO2 & R$^3$LIVE & Ctrl-VIO \\
        \midrule
        JT-O1   & Outdoor & 109 & E2E  & \textbf{0.02} & 0.15 & \textbf{0.02} & 704.59 & 42.27 & 7.80 \\
        JT-O2   & Outdoor & 94  & E2E  & \textbf{0.04} & 0.16 & \textbf{0.04} & 0.12 & 1.42 & 443.51 \\
        JT-O3   & Outdoor & 70  & E2E  & \textbf{0.03} & \textbf{0.03} & \textbf{0.03} & 0.05 & 2.11 & 25.16 \\
        JT-I1   & Indoor  & 99  & E2E  & \textbf{0.02} & 0.05 & \textbf{0.02} & 0.04 & 0.12 & 2.34 \\
        JT-M1   & Indoor  & 55  & RMSE & 0.042 & \textbf{0.041} & 0.044 & 0.043 & 0.049 & 0.157 \\
        JT-M2   & Indoor  & 56  & RMSE & \textbf{0.043} & 0.054 & \textbf{0.043} & 0.169 & 0.247 & 0.255 \\
        AVIA-I1 & Indoor  & 35  & E2E  & 0.04 & \textbf{0.03} & 0.04 & 0.04 & 0.06 & 0.15 \\
        AVIA-O1 & Outdoor & 164 & E2E  & \textbf{0.05} & 0.08 & \textbf{0.05} & 57.86 & 19.61 & 9.36 \\
        AVIA-O2 & Outdoor & 54  & E2E  & \textbf{0.05} & 2.31 & 1.26 & 5.68 & 4.26 & 16.13 \\
        \midrule
        \yellowtext{GRI-1} & \yellowtext{GrandTour} & \yellowtext{455} & \yellowtext{RMSE} & \yellowtext{\textbf{0.076}} & \yellowtext{0.079} & \yellowtext{0.080} & \yellowtext{0.087} & \yellowtext{0.095} & \yellowtext{48.742} \\
        \yellowtext{SPX-3} & \yellowtext{GrandTour} & \yellowtext{213} & \yellowtext{RMSE} & \yellowtext{0.019} & \yellowtext{\textbf{0.018}} & \yellowtext{0.019} & \yellowtext{\textbf{0.018}} & \yellowtext{0.021} & \yellowtext{2.164} \\
        \yellowtext{SBB-1} & \yellowtext{GrandTour} & \yellowtext{502} & \yellowtext{RMSE} & \yellowtext{\textbf{0.068}} & \yellowtext{0.071} & \yellowtext{0.069} & \yellowtext{0.091} & \yellowtext{0.096} & \yellowtext{25.478} \\
        \yellowtext{HEAP-1} & \yellowtext{GrandTour} & \yellowtext{290} & \yellowtext{RMSE} & \yellowtext{0.026} & \yellowtext{0.026} & \yellowtext{0.027} & \yellowtext{\textbf{0.021}} & \yellowtext{0.026} & \yellowtext{1.023} \\
        \bottomrule
    \end{tabular}}
    \vspace{4pt}
    \begin{minipage}{\textwidth}
    \footnotesize
    \raggedright
    \textsuperscript{1} All error values are reported in meters. \par
    \end{minipage}
    \vspace{-14pt}
\end{table*}

\section{Experiments}
\label{sec:experiments}

\subsection{Experimental Setup}
\label{sec:exp_setup}

All experiments run on Ubuntu 20.04 with an Intel i7-10875H CPU, with the same number of cores bound to each process and locked to the maximum frequency. We compare FAST-LIVO-RS against FAST-LIVO2~\cite{zheng2024livo2} (GS assuming direct LIVO), R$^3$LIVE~\cite{lin2021r3live} (direct LIVO with a denser visual backend), and Ctrl-VIO~\cite{lang2022ctrlvio} (RS aware B-spline \yellowtext{VIO). Table}~\ref{tab:rs_parameters} lists the configuration used in every reported experiment. We define two component-level ablations. The single-segment variant removes adaptive state augmentation and retains only the two poses at the beginning and end of the readout interval. The map-interaction variant removes RS-consistent visual map management.

\yellowtext{We additionally evaluate four sequences from the GrandTour dataset}~\cite{tuna2026grandtour}
\yellowtext{using the forward-facing TierIV C1 HDR rolling-shutter camera, Hesai XT-32 LiDAR, and STIM320 IMU. Ctrl-VIO uses the same camera and IMU without LiDAR input.}

\begin{figure}[t]
    \centering
    \includegraphics[width=0.99\columnwidth]{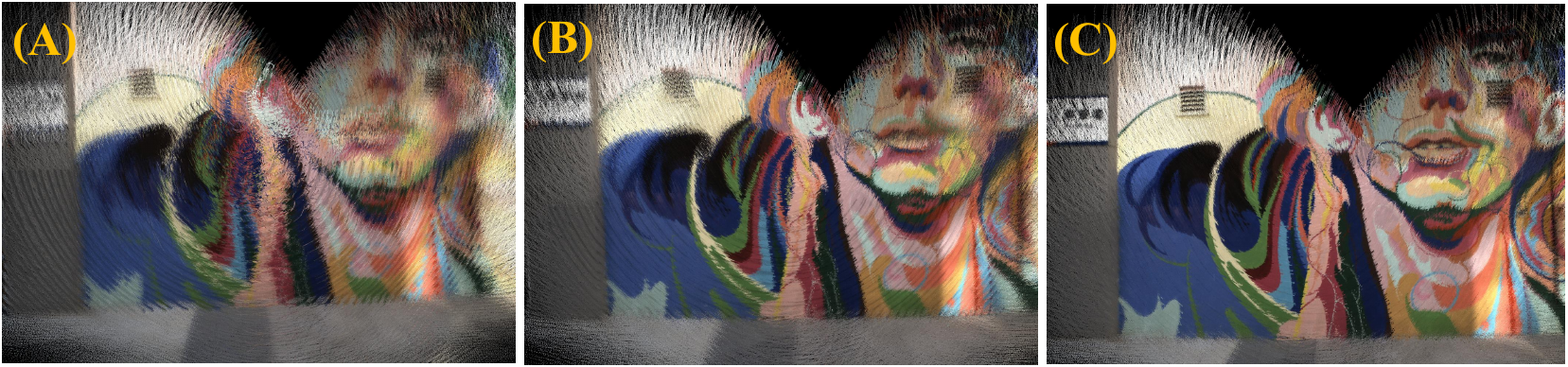}
    \caption{Ablation visualization of adaptive state augmentation. From left to right: FAST-LIVO2, FAST-LIVO-RS without adaptive state augmentation, and full FAST-LIVO-RS.}
    \label{fig:adaptive_clone_ablation}
\end{figure}

\subsection{Quantitative Accuracy Evaluation}
\label{sec:quantitative_eval}

Table~\ref{tab:main_results} summarizes the trajectory accuracy on the self-collected and GrandTour datasets.
On the self-collected data, FAST-LIVO-RS matches or outperforms all external baselines across the evaluated sequences. Because every sequence was collected with a handheld rig, gait-induced jolts can create brief rapid rotations. JT-O1 and AVIA-O1 also contain extended LiDAR-degenerate intervals. Under these conditions, weak LiDAR geometry increases the estimator's reliance on visual constraints, while the rapid rotations amplify row-dependent photometric mismatch under FAST-LIVO2's GS assumption. The combined effect causes FAST-LIVO2 to suffer severe drift on both sequences. Adaptive state augmentation is particularly effective on AVIA-O2, which was collected under deliberately vigorous shaking of the handheld rig. The resulting rapid intra-frame motion makes additional temporal support especially important. The map-interaction ablation remains close to the full method on most sequences, but exhibits its clearest degradation on AVIA-O2, demonstrating the importance of row-specific map geometry when RS effects dominate the visual constraint. \yellowtext{Ctrl-VIO, which has no LiDAR geometric constraint, accumulates pronounced drift over the long GrandTour sequences and therefore exhibits large trajectory errors.}

Fig.~\ref{fig:colorization_comparison} provides qualitative evidence of improved row-consistent visual-map alignment. Even when FAST-LIVO2 does not show a large E2E drift, its GS assuming treatment of RS images can still yield inconsistent visual-map geometry, producing smeared wall text, blurred mural boundaries, and ghosting near high-contrast display edges. FAST-LIVO-RS estimates a row-consistent continuous trajectory, so the colored map preserves sharper text strokes, cleaner object boundaries, and more coherent texture on vertical structures. The same behavior is also visible in the accompanying video.

Fig.~\ref{fig:adaptive_clone_ablation} visualizes the effect of adaptive state augmentation beyond the scalar trajectory metrics. Even when the E2E or RMSE values are close, the single-segment model still leaves color bleeding around mural contours and high-contrast text, because one linear segment cannot fully represent the intra-frame rotational variation over the readout interval. Adaptive state augmentation produces a sharper and more coherent colored point cloud, showing that the extra temporal resolution improves the consistency of the estimated continuous trajectory.

\begin{table}[t]
    \centering
    \caption{Average per-frame processing time (ms).}
    \label{tab:runtime}
    \resizebox{0.4\textwidth}{!}{%
    \begin{tabular}{l c c c c}
        \toprule
        Platform & Ours & FAST-LIVO2 & Ctrl-VIO & R$^3$LIVE \\
        \midrule
        JT & 26.23 & \textbf{24.07} & 162.87 & 83.13 \\
        AVIA & 24.73 & \textbf{22.43} & 160.18 & 76.62 \\
        \bottomrule
    \end{tabular}}
\end{table}
\subsection{Runtime Analysis}
\label{sec:runtime_eval}
Table~\ref{tab:runtime} reports the per-frame processing time averaged separately over the six JT sequences and the three AVIA sequences. FAST-LIVO-RS remains close to FAST-LIVO2 on both platforms, with a modest overhead caused mainly by maintaining the augmented clone states and performing fixed-point row projection during the visual update. These operations are bounded by the clone budget and remain within the same sequential ESIKF pipeline. Ctrl-VIO is substantially slower because its keyframe-based sliding-window B-spline optimization requires repeated control-point marginalization~\cite{lang2022ctrlvio}.

\subsection{\texorpdfstring{RS and GS Performance Comparison}{Controlled Evaluation with RS and GS Cameras}}
\label{sec:controlled_rs_gs}

We conduct an independent controlled comparison using synchronized RS and GS cameras on the six JT recordings. Before evaluation, images from the RS camera are downsampled to the resolution of the GS camera. The three configurations are FAST-LIVO-RS using images from the RS camera, FAST-LIVO2 using the same RS images, and FAST-LIVO2 using images from the synchronized GS camera. For reconstruction, all three configurations use 3DGUT. FAST-LIVO-RS+RS provides the estimated row-time trajectory; for FAST-LIVO2+RS and FAST-LIVO2+GS, we assign each estimated single-frame pose to all image rows of the corresponding frame. For every valid reconstruction, we report peak signal-to-noise ratio (PSNR) and structural similarity index measure (SSIM).
\begin{figure}[t]
    \centering
    \includegraphics[width=0.99\columnwidth]{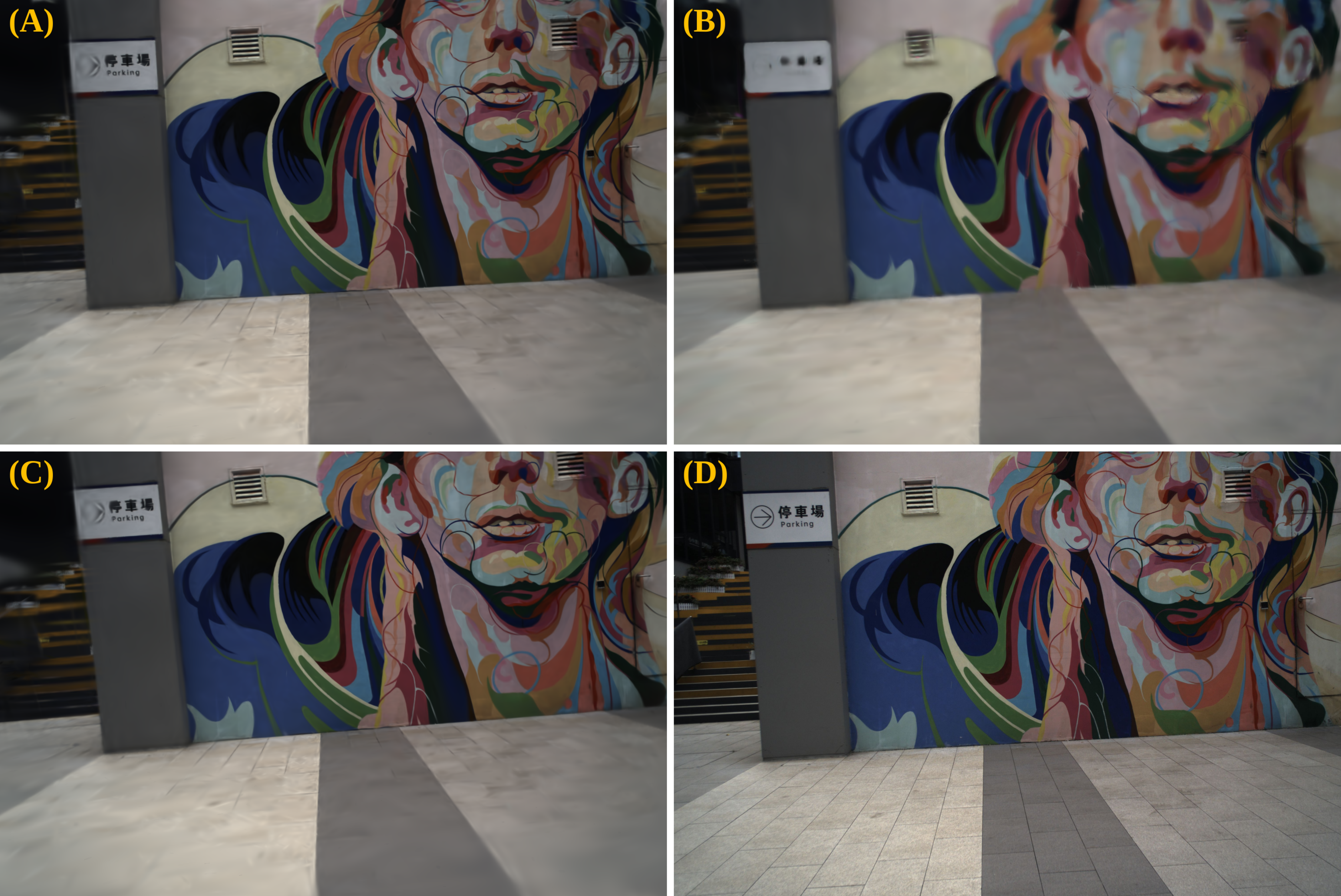}
    \caption{Reconstruction comparison under the synchronized RS/GS setup: (A) FAST-LIVO-RS+RS with 3DGUT, (B) FAST-LIVO2+RS with 3DGUT, (C) FAST-LIVO2+GS with 3DGUT, and (D) the corresponding reference image. The synchronized RS and GS test views are temporally corresponding but not strictly pose-aligned because the cameras have different optical centers.}
    \label{fig:3dgs_comparison}
    \vspace{6pt}
\end{figure}
\begin{table}[t]
    \centering
    \caption{Odometry and reconstruction results.}
    \label{tab:controlled_rs_gs}
    \scriptsize
    \setlength{\tabcolsep}{2pt}
    \renewcommand{\arraystretch}{0.92}
    \resizebox{\columnwidth}{!}{%
    \begin{tabular}{@{}llccc@{}}
        \toprule
        Sequence & Metric
        & \shortstack{FAST-LIVO-RS\\+RS}
        & \shortstack{FAST-LIVO2\\+RS}
        & \shortstack{FAST-LIVO2\\+GS} \\
        \midrule
        JT-O1 & E2E [m] & \textbf{0.02} & 186.51 & 0.03 \\
        & PSNR [dB] & 25.12 & -- & \textbf{26.06} \\
        & SSIM & \textbf{0.778} & -- & 0.770 \\
        \addlinespace[1pt]
        JT-O2 & E2E [m] & 0.06 & 0.14 & \textbf{0.05} \\
        & PSNR [dB] & 24.21 & 21.70 & \textbf{25.56} \\
        & SSIM & 0.742 & 0.719 & \textbf{0.755} \\
        \addlinespace[1pt]
        JT-O3 & E2E [m] & \textbf{0.03} & 0.05 & \textbf{0.03} \\
        & PSNR [dB] & 29.43 & 28.82 & \textbf{29.67} \\
        & SSIM & \textbf{0.857} & 0.829 & 0.850 \\
        \addlinespace[1pt]
        JT-I1 & E2E [m] & \textbf{0.02} & 0.03 & \textbf{0.02} \\
        & PSNR [dB] & 31.61 & 29.68 & \textbf{31.96} \\
        & SSIM & \textbf{0.896} & 0.845 & 0.882 \\
        \addlinespace[1pt]
        JT-M1 & RMSE [m] & 0.042 & 0.044 & \textbf{0.039} \\
        & PSNR [dB] & \textbf{26.51} & 24.65 & 26.06 \\
        & SSIM & \textbf{0.862} & 0.810 & 0.857 \\
        \addlinespace[1pt]
        JT-M2 & RMSE [m] & \textbf{0.041} & 0.153 & \textbf{0.041} \\
        & PSNR [dB] & 23.90 & 20.49 & \textbf{24.58} \\
        & SSIM & 0.855 & 0.809 & \textbf{0.865} \\
        \midrule
        Valid average & PSNR [dB] & 26.80 & 25.07 & \textbf{27.32} \\
        & SSIM & \textbf{0.832} & 0.802 & 0.830 \\
        \bottomrule
    \end{tabular}}
    \vspace{6pt}
    \begin{minipage}{\columnwidth}
    \scriptsize\raggedright
    \vspace{2pt}
    ``--'' marks an invalid or inapplicable entry.\par
    \end{minipage}
    \vspace{-16pt}
\end{table}

Under the common 3DGUT reconstruction backend, FAST-LIVO-RS+RS achieves aggregate reconstruction quality close to FAST-LIVO2+GS and higher than FAST-LIVO2+RS. Because the configurations use different trajectory inputs and camera shutter conditions, these metrics characterize the complete odometry--reconstruction pipelines rather than reconstruction quality alone. The odometry results also show the same overall robustness trend: FAST-LIVO-RS+RS approaches the GS-camera configuration while remaining more robust than FAST-LIVO2+RS under pronounced RS effects.

Fig.~\ref{fig:3dgs_comparison} compares the three reconstruction results under the synchronized RS/GS setup, with the corresponding reference image shown in (D). FAST-LIVO2+RS with 3DGUT exhibits pronounced blur along the mural contours and parking sign, whereas FAST-LIVO-RS+RS with the same 3DGUT backend preserves sharper boundaries and more legible signage, approaching the synchronized FAST-LIVO2+GS reference. Consistent with Table~\ref{tab:controlled_rs_gs}, this is a qualitative comparison of the complete odometry--reconstruction pipelines rather than an isolated renderer test.

\begin{figure}[!t]
    \centering
    \includegraphics[width=0.94\columnwidth]{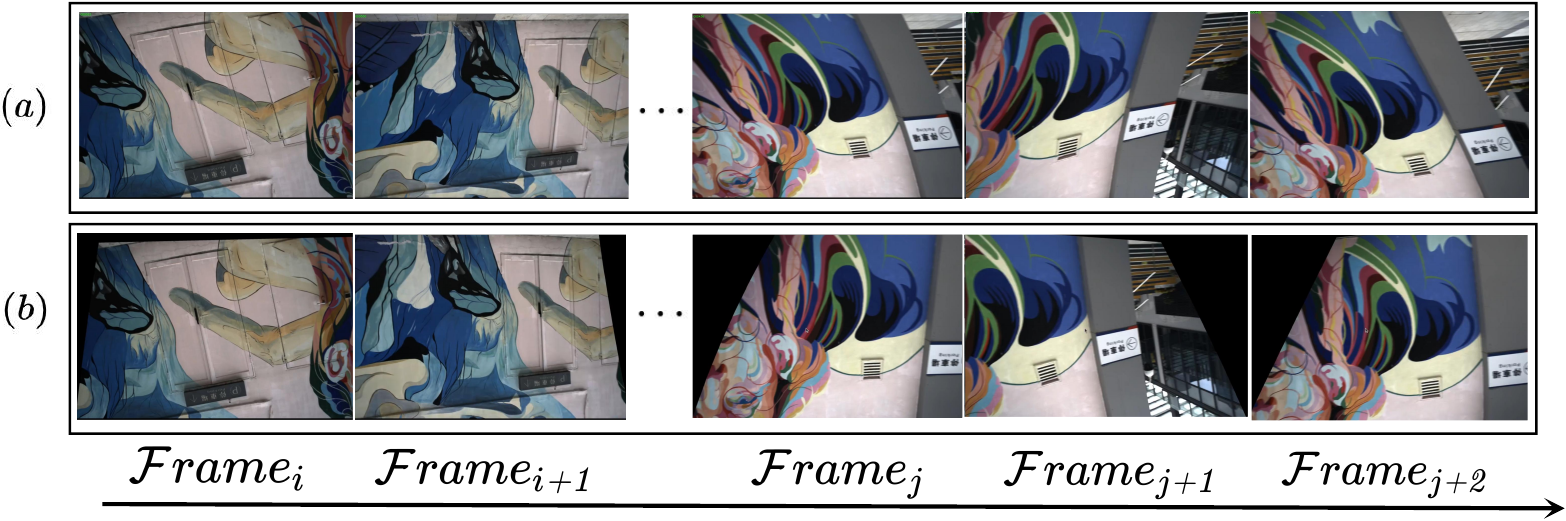}
    \caption{RS image undistortion examples. The top row shows raw RS frames, and the bottom row shows undistorted outputs.}
    \label{fig:undistortion}
\end{figure}

\subsection{\texorpdfstring{Image Undistortion}{Image Undistortion}}
\label{sec:undistortion_eval}

Under aggressive rotational motion, rolling-shutter distortion produces visible shear and skew in vertical structures and inconsistent geometry across image rows. Fig.~\ref{fig:undistortion} shows that the output produced by the row homography~\eqref{eq:H_undist} removes the dominant inter-row rotational distortion and yields a GS-equivalent image associated with the end pose. The black regions in the lower row are invalid pixels: after fixed-point remapping, their source coordinates lie outside the valid RS image area, so no measured pixel can be sampled there. 
\section{Conclusion}
This paper presents FAST-LIVO-RS, a rolling-shutter aware direct LiDAR-inertial-visual odometry system that combines adaptive state augmentation for RS motion, row-consistent photometric updates, and RS-consistent visual map management within a sequential ESIKF. Experiments on self-collected and public datasets, synchronized RS/GS comparisons, and downstream applications demonstrate improved robustness and trajectory consistency under rolling-shutter distortion while retaining real-time efficiency. Future work will investigate online timing calibration and more dynamic environments.

\bibliography{paper}

\end{document}